\documentclass{article} %
\usepackage{iclr2025_conference,times}

\usepackage{amsmath,amsfonts,bm}

\def\eqref#1{equation~\ref{#1}}

\def\1{\bm{1}}

\DeclareMathAlphabet{\mathsfit}{\encodingdefault}{\sfdefault}{m}{sl}
\SetMathAlphabet{\mathsfit}{bold}{\encodingdefault}{\sfdefault}{bx}{n}

\usepackage{hyperref}
\usepackage{url}
\usepackage{booktabs}
\usepackage{graphicx}
\usepackage{xcolor}
\usepackage{multirow}
\usepackage[capitalize]{cleveref}

\newcommand{\rot}[1]{\rotatebox[origin=c]{90}{#1}}

\newcommand{\eg}{\textit{e.g.}}
\newcommand{\ie}{\textit{i.e.}}

\definecolor{darkergreen}{RGB}{21, 152, 56}
\definecolor{red2}{RGB}{252, 54, 65}

\newcommand{\pp}[1]{\textcolor{darkergreen}{\small{\textbf{(+#1)}}}}
\newcommand{\np}[1]{\textcolor{red2}{\small{\textbf{(-#1)}}}}

\title{LongProLIP: A Probabilistic Vision-Language Model with Long Context Text}

\author{Sanghyuk Chun \qquad Sangdoo Yun\\
NAVER AI Lab}

\iclrfinalcopy %
\begin{document}

\maketitle

\begin{abstract}
Recently, Probabilistic Language-Image Pre-Training (ProLIP) has been proposed to tackle the multiplicity issue of vision-language (VL) tasks. Despite their success in probabilistic representation learning at a scale, the ProLIP models cannot handle long context texts longer than 64 context length, which limits their ability to capture rich contextual information from longer text sequences. To address this issue, this paper proposes a fine-tuning strategy for ProLIP to accept longer texts, \eg, 256 text tokens. Experimental results on Urban-1k and the DataComp evaluation suite show that the proposed LongProLIP recipe can improve understanding of long contexts while minimizing the negative effect of fine-tuning. We also observe a trade-off between the long context understanding (measured by Urban-1k) and general zero-shot capability (measured by evaluation datasets by DataComp). Code is available at \url{https://github.com/naver-ai/prolip}.
\end{abstract}

\section{Introduction}

\begin{figure}[t]
    \centering
    \includegraphics[width=.7\linewidth]{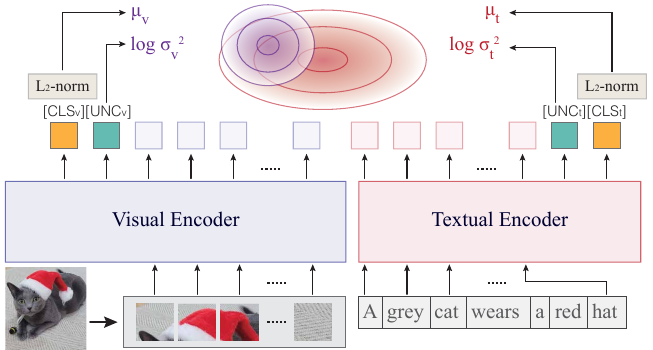}
    \caption{\small {\bf Overview of ProLIP.} ProLIP consists of an image encoder and a text encoder. The mean and variance are estimated by \texttt{[CLS]} token and \texttt{[UNC]} token, respectively. Note that the original CLIP text encoder does not use \texttt{[CLS]} token, but the ProLIP text encoder uses the last additional two tokens for \texttt{[CLS]} and \texttt{[UNC]} tokens. ProLIP is trained with probabilistic objective functions, such as probabilistic pairwise contrastive loss, inclusion loss, and variational information bottleneck loss.}
    \label{fig:main}
\end{figure}

Probabilistic vision-language models (PrVLMs) \citep{pcme,ji2023map,upadhyay2023probvlm,pcmepp,prolip} aim to tackle the multiplicity problem of VL tasks, \eg, an image can be described by multiple captions and vice versa. Recently, \citet{prolip} proposed ProLIP, the first PrVLM pre-trained on billion-scale image-text pairs (trained on DataComp 1B \citep{datacomp} with 12.8B training seen samples). As shown in \cref{fig:main}, ProLIP estimates both the mean and variance vectors of a Gaussian probabilistic embedding using the \texttt{[CLS]} token and the \text{[UNC]} token, respectively. By using the uncertainty estimated by the extracted variance, \citet{prolip} showed the advantages of uncertainty estimation in VL tasks. For example, ProLIP shows that shorter texts are more uncertain and more general inputs including specific ones which can help the understanding of the given dataset. Also, ProLIP can improve the prompt selection for zero-shot classification tasks and image traversal tasks with uncertainty.

Despite its success on capturing the ambiguity of VL tasks with short captions, ProLIP has a limitation at its text context length; the original ProLIP text encoder only takes at most 64 text tokens. However, in practice, text contexts often exceed 64 tokens. For example, as shown by LongCLIP \citep{longclip}, a text prompt for a text-to-image generation task can be longer than 64 tokens to describe very fine-grained details. As another example, the captions generated by large vision-language models, such as LLaVA \citep{llava}, usually have very long text lengths (\eg, larger than 100 tokens). For practical usability, we argue that 64 token length might not be sufficient.

In this paper, we propose \textbf{LongProLIP}, an extension of ProLIP by taking longer text context length, (\ie, 256). Our approach is based on LongCLIP \citep{longclip}, but we provide a more detailed study of the training dataset. Also, we observe that directly applying the LongCLIP recipe to ProLIP often leads to a significant performance drop (\eg, ImageNet zero-shot accuracy becomes 3.7\% from 67.8\%), especially when the base pre-trained model is not sufficiently strong.
More specifically, we observe a trade-off between the long context understanding (measured by Urban-1k) and general zero-shot capability (measured by ImageNet or the average of 38 zero-shot evaluation datasets by DataComp).
When we focus on the long context understanding, our LongProLIP model achieved the state-of-the-art performance on the Urban-1k dataset, proposed to measure the long context understanding ability of VLM \citep{longclip}; while LongCLIP ViT-L/14 achieves 82.7\% I2T R@1 and 86.1\% T2I R@1, our LongProLIP ViT-B/16 achieves 90.8\% and 91.8\%, respectively.
The LongProLIP weights will be available at the HuggingFace hub upon acceptance.

\section{Results}

\paragraph{Pre-trained models.}
We use the official pre-trained weights provided by \citet{prolip}, including two ProLIP ViT-B/16 models trained with 12.8B seen samples and 1.28B seen samples, trained on DataComp 1B. To compare LongProLIP and LongCLIP, we compare the ProLIP model trained with 1.28B seen samples to the CLIP model trained on the same number of seen samples.

\paragraph{Fine-tuning dataset.}
Following LongCLIP \citep{longclip}, we fine-tune the models on the ShareGPT4V dataset \citep{sharegpt4v} which contains 1.2M image-text pairs with highly descriptive captions. As shown in \cref{appendix:sec:example_sharegpt}, these captions contain very long text tokens, such as more than 150.
However, our experiments show that using only ShareGPT4V for fine-tuning often leads to a severe performance drop in general zero-shot classification, such as ImageNet. To tackle the issue, we employ a high-quality filtered dataset, the filtered DataComp medium with HYPE \citep{hype} and Data Filtering Network (DFN) \citep{dfn}.

\paragraph{Optimization and hyperparameters.}
We fine-tune the models with the learning rate of 1e-06, the weight decay of 0.01, and the batch size of 8,192. The other optimization hyperparameters are the same as \citet{prolip}.
Also, following LongCLIP \citep{longclip}, we interpolate the 66-dimensional positional embedding (64 context length + 1 \texttt{[CLS]} token + 1 \texttt{[UNC]} token), while the first 20 tokens (preserving the pre-trained textual knowledge) and the last 2 tokens (preserving the information of \texttt{[CLS]} and \texttt{[UNC]} tokens) are fixed during the interpolation.

\paragraph{Evaluation dataset.}
We mainly compare the models on the Urban-1k dataset proposed by \citet{longclip}, which measures the long context text understanding ability. We additionall measure the general zero-shot capability by ImageNet \citep{imagenet}, ECCC Caption \citep{eccv_caption} (an extended version of COCO Caption \citep{coco_caption}), and 38 zero-shot evaluation suite by DataComp \citep{datacomp}. We list the full dataset of 38 tasks in \cref{appendix:sec:evaluation_dataset}.

\paragraph{Comparison with LongCLIP.}
\cref{tab:longprolip_vs_longclip} shows the comparison of LongCLIP and LongProLIP in Urban-1k and ImageNet, where each measures the long context text understanding ability and the general zero-shot classification ability.
For a fair comparison, we use both CLIP and ProLIP models pre-trained from DataComp 1B with 1.28B training seen samples.

In the table, we observe two main findings. First, ProLIP and LongProLIP show a better long context text understanding compared to CLIP and LongCLIP. We presume that this is due to uncertainty-aware modeling, which can capture ``specificity'' or ``general'' concepts in the given texts.
Second, while LongCLIP fine-tuned on ShareGPT4V does not show a significant performance drop in ImageNet zero-shot classification (IN ZSC), ProLIP fine-tuned solely on ShareGPT4V shows a severe performance drop in the same setting (67.8 $\rightarrow$ 3.7). Interestingly, as we will show in the latter experiments, if we use a stronger ProLIP backbone, the performance drop is not as significant as \cref{tab:longprolip_vs_longclip}.
We presume that this is because the uncertainty modeling of the ProLIP with 1.28B seen sample model is not sufficient, therefore, during the fine-tuning with probabilistic objectives, the learned probabilistic space is largely changed.

We additionally fine-tune ProLIP with high-quality short image-text pairs, \ie, the combination of HYPE medium \citep{hype} and DFN medium \citep{dfn}. In the table, we observe that while the ImageNet performance drop is neglectable (-0.9\%), the average long context understanding ability is improved after the fine-tuning (+3.8\%). From this observation, we propose to use the LongProLIP model fine-tuned with ShareGPT4V + HYPE + DFN for a general purpose, and use the LongProLIP model fine-tuned solely with ShareGPT4V for a long context-specific purpose.

\begin{table}[t]
    \centering
    \small
    \caption{\small {\bf Comparison of LongCLIP and LongProLIP.} We use CLIP and ProLIP models trained on DataComp 1B with 1.28B seen samples. Unlike CLIP, directly applying the LongCLIP recipe to ProLIP leads to a severe performance drop to zero-shot classification (ZSC), such as ImageNet ZSC.}
    \label{tab:longprolip_vs_longclip}
    \vspace{.5em}
    \begin{tabular}{llllll}
    \toprule
    Model & Fine-tuning datasets & Urban-1k I2T & Urban-1k T2I & Urban-1k (Avg) & ImageNet \\
    \midrule
    \multirow{2}{*}{CLIP} & - & 41.9 & 44.8 & 43.4 & 67.2 \\
    & ShareGPT4V only & 83.1 \pp{41.2} & 80.1 \pp{35.3} & 81.6 \pp{38.2} & 63.3 \np{3.9} \\ \midrule
    \multirow{3}{*}{ProLIP}  & - & 56.2 & 54.7 & 55.5 & 67.8 \\
    & ShareGPT4V only & 84.9 \pp{28.7} & 86.6 \pp{31.9} & 85.8 \pp{30.3} & 3.7 \np{64.1} \\
    & ShareGPT4V + HYPE + DFN & 55.7 \np{0.5} & 62.9 \pp{8.2} & 59.3 \pp{3.8} & 66.9 \np{0.9} \\
    \bottomrule
    \end{tabular}
\end{table}

\begin{table}[t]
    \centering
    \small
    \caption{\small {\bf LongProLIP results by different fine-tuning datasets.} $N$ denotes the number of the seen samples during fine-tuning.
    Note that Urban-1k Avg results of LongCLIP are 79.2 for ViT-B/16 and 84.4 for ViT-L/14, respectively.
    We additionally report ECCV Caption mAP@R and the average score of the 38 tasks of the DataComp evaluation suite. The full results are shown in \cref{appendix:tab:overview} and \cref{appendix:tab:full}
    }
    \label{tab:longprolip_main}
    \vspace{.5em}
    \begin{tabular}{llllll}
    \toprule
    Fine-tuning datasets & $N$ & Urban-1k (Avg) & EC mAP (Avg) & ImageNet & 38 Tasks (Avg)   \\
    \midrule
    - &  - & 65.4 & 34.1 & 74.6 & 63.3        \\ \midrule
    ShareGPT4V only & 24M & 87.2 \pp{21.8} & 35.7 \pp{1.6} & 70.8 \np{3.8} & 60.5 \np{2.8} \\
    ShareGPT4V only & 128M & 91.3 \pp{25.9} & 33.4 \np{0.7} & 69.5 \np{5.1} & 58.7 \np{4.6} \\
    ShareGPT4V + HYPE + DFN & 128M & 77.5 \pp{12.1} & 34.6 \pp{0.5} & 74.6 \pp{0.0} & 63.3 \pp{0} \\
    \bottomrule
    \end{tabular}
\end{table}

\paragraph{Main results.}
In \cref{tab:longprolip_main}, we show the LongProLIP results with a stronger ProLIP backbone, \ie, ViT-B/16 with 12.8B seen samples. We observe similar results to \cref{tab:longprolip_vs_longclip}, except the performance drop is not as significant as \cref{tab:longprolip_vs_longclip}.
We observe that when we fine-tune ProLIP solely on ShareGPT4V with more iterations (\ie, 24M seen samples vs. 128M seen samples), the Urban-1k result is improved but the other performances are all dropped. On the other hand, when we use ShareGPT4V and HYPE + DFN for fine-tuning, the overall performance has been improved, including retrieval and zero-shot capability.
Specifically, in the full results in \cref{appendix:tab:overview} and \cref{appendix:tab:full}, we can observe that the retrieval performances (\eg, COCO Caption \citep{coco_caption}, WinoGAViL \citep{winogavil}, and ECCV Caption \citep{eccv_caption}) are largely improved with the LongProLIP fine-tuning. While ShareGPT4V only fine-tuning often leads to severe performance drops, the mixed fine-tuning strategy seems to prevent the performance drops (\eg, the KITTI task \citep{kitti} shows -16.32\% and -10.69\% drops for 24M and 128M ShareGPT4V fine-tuning, but the mixed strategy shows +3.17\% performance improvement compared to the original performance).

\paragraph{Conclusion and future work.}
This paper proposes LongProLIP, an extension of ProLIP with a longer text context (64 $\rightarrow$ 256). We observe that fine-tuning solely with the ShareGPT4V dataset can lead to a significant performance drop for general zero-shot tasks.
The future work would include analyses of why ProLIP is more sensitive than CLIP in a long text fine-tuning setting, and improving our understanding of VL tasks with long captions (\eg, text-to-image generation or highly descriptive captioning) using the newly obtained long context model.

\bibliography{iclr2025_conference}
\bibliographystyle{iclr2025_conference}

\clearpage
\appendix
\numberwithin{equation}{section}
\numberwithin{figure}{section}
\numberwithin{table}{section}

\section*{Appendix}

\section{Examples of ShareGPT4V caption}
\label{appendix:sec:example_sharegpt}

Here, we provide examples of ShareGPT4V captions \citep{sharegpt4v}. These captions are highly descriptive and have very long context (\eg, more than 180 tokens).

{\small
\begin{itemize}
\item This image captures a serene moment on a sandy beach. At the center of the frame, a golden retriever is sitting comfortably inside a heart drawn in the sand. The dog's pink tongue is sticking out, adding a playful touch to the scene. The heart, drawn with a stick or similar object, is large and occupies most of the foreground of the photo. The sandy texture of the beach contrasts with the smooth lines of the heart. The beach itself is empty, creating a sense of tranquility. In the background, you can see the shoreline and the water, which adds depth to the image. Further in the distance, there's a line of trees and houses, providing a hint of civilization. The sky above is overcast, casting a soft light over the entire scene and enhancing the peaceful atmosphere. Despite the absence of bright sunlight, the image exudes warmth, largely due to the presence of the golden retriever at its heart. Overall, this image beautifully combines elements of nature with a symbol of love and companionship.
\item In the image, a young man is seen comfortably reclining on a vibrant purple couch. He is wearing a stylish black jacket and a pair of glasses that reflect the light from the two laptops resting on his chest. The laptops, both white in color, are adorned with an assortment of colorful stickers, adding a personal touch to the devices. The man appears to be deeply engrossed in his work or perhaps enjoying some digital entertainment. The couch is positioned on a gray carpeted floor which extends out into the room, providing a soft contrast to the bold color of the couch. The beige wall in the background adds a neutral backdrop to the scene, allowing the focus to remain on the man and his activities. The image captures a moment of modern life, where technology is an integral part of our daily routines. It's a snapshot of digital age comfort and productivity.
\item This image captures a dynamic scene on a snowy mountain. At the center of the frame, a skier dressed in a black jacket and pants is in action. The skier, equipped with a helmet and goggles for safety, skillfully maneuvers two ski poles to aid in their descent. The skier is captured mid-turn, their body leaning into the curve as they carve a path down the slope. This action kicks up a spray of snow, creating a dramatic effect against the serene backdrop. The location is a mountain blanketed in snow, its surface untouched except for the trail left by the skier. The mountain's incline suggests a steep descent, adding to the thrill of the scene. In the background, a forest of trees stands tall. Their branches are laden with snow, painting a picturesque winter landscape. The trees appear dense and extend far into the distance, providing a stark contrast to the open space being navigated by the skier. Overall, this image encapsulates the exhilarating sport of skiing against the tranquil beauty of a snowy mountain landscape.
\end{itemize}
}

\section{Datacomp evaluation suite details}
\label{appendix:sec:evaluation_dataset}

The DataComp evaluate suite contains 38 zero-shot tasks: ImageNet \citep{imagenet}, 6 ImageNet distribution shifts robust benchmarks, including ImageNet-A, ImageNet-O \citep{imagenet_a}, ImageNet-R \citep{imagenet_r}, ImageNet v2 \citep{imagenet_v2}, ImageNet-Sketch \citep{imagenet_sketch} and ObjectNet \citep{objectnet}, 13 VTAB task \citep{vtab}, including Caltech-101 \citep{caltech101}, CIFAR-100 \citep{cifar}, CLEVR Counts, CLEVR Distance \citep{clevr}, Describable Textures \citep{dtd}, EuroSAT \citep{eurosat}, KITTI Vehicle Distance \citep{kitti}, Oxford Flowers-102 \citep{flowers102}, Oxford-IIIT Pet \citep{pets}, PatchCamelyon \citep{pcam}, RESISC45 \citep{resisc45}, SVHN \citep{svhn} and SUN397 \citep{sun}, and 3 retrieval tasks, including Flickr \citep{flickr30k}, MS-COCO Caption \citep{coco_caption} and WinoGAViL \citep{winogavil}.
There are also 13 additional tasks, such as CIFAR-10, Country211 \citep{radford2021clip}, FGVC Aircraft \citep{fgvc}, Food-101 \citep{food101}, GTSRB \citep{gtsrb}, MNIST \citep{mnist}, Pascal VOC \citep{voc}, Rendered SST2 \citep{radford2021clip}, STL-10 \citep{stl10}, iWildCam \citep{iwildcam}, FMoW \citep{fmow}, Dollar Street \citep{dollarstreet}, and GeoDE \citep{geode}.

\section{Full results}
\label{appendix:sec:full_results}

\cref{appendix:tab:overview} shows the additional information from \cref{tab:longprolip_main}. We can observe that ShareGPT4V fine-tuning for ProLIP may harm the zero-shot ability (\eg, ImageNet, ImageNet robustness benchmark and VTAB), but it can be beneficial to retrieval tasks, such as Flickr, MS-COCO Caption, WinoGAViL, Urban-1k and ECCV Caption.

\cref{appendix:tab:full} shows the full results of the DataComp evaluation suite (38 tasks). Similar to \cref{appendix:tab:overview}, LongProLIP fine-tuning specifically improves retrieval performances.

\begin{table}[t]
    \centering
    \small
    \caption{\small {\bf Zero-shot classification full results.} IN dist. denotes the average of 6 ImageNet distribution shift robust benchmarks and VTAB contains 13 zero-shot tasks. Retrieval is the average of three retrieval tasks. ``S24M'' denotes that the model is fined-tuned on ShareGPT4V with 24M seen samples; S128M is defined similarly (128M seen samples). SHD128M denotes that the model is fine-tuned on ShareGPT4V, HYPE, and DFN with 128M seen samples. Each four row corresponds to each row in \cref{tab:longprolip_main}.}
    \label{appendix:tab:overview}
    \vspace{.5em}
    \setlength{\tabcolsep}{2pt}
    \resizebox{\textwidth}{!}{
    \begin{tabular}{llllllll}
    \toprule
    & IN dist. & VTAB & Retrieval & Urban I2T R@1 & Urban T2I R@1 & EC mAP@R I2T & EC mAP@R T2I \\
    \midrule
    ProLIP (pre-trained) & 63.0 & 63.7 & 59.6 & 67.5 & 63.3 & 28.9 & 39.2 \\
    ProLIP (S24M) & 59.1  \np{-3.90} &  60.9  \np{-2.80} &  61.3  \pp{1.70} &  86.9 \pp{19.4}  &  87.4 \pp{24.1}  &  30.5  \pp{1.60} &  41.0 \pp{1.80} \\
    ProLIP (S128M) & 58.1  \np{-4.90} &  58.4  \np{-5.30} &  59.6  \pp{0.00} &  90.8 \pp{23.3}  &  91.8 \pp{28.5}  &  29.6  \pp{0.70} &  37.2 \np{-2.00} \\
    ProLIP (SHD128M) & 62.5  \np{-0.50} &  63.0  \np{-0.70} &  61.9  \pp{2.30} &  75.8 \pp{8.3}  &  79.1 \pp{15.8}  &  29.2  \pp{0.30} &  40.1 \pp{0.90} \\
    \bottomrule
    \end{tabular}
    }
\end{table}

\begin{table}[t]
    \centering
    \small
    \caption{\small {\bf Zero-shot classification full results.} The detail is same as \cref{appendix:tab:overview}.}
    \label{appendix:tab:full}
    \vspace{.5em}
    \setlength{\tabcolsep}{2pt}
    \resizebox{\textwidth}{!}{
    \begin{tabular}{l*{11}{c}}
    \toprule
    & \rot{ImageNet 1k} & \rot{CIFAR-10} & \rot{CIFAR-100} & \rot{CLEVR Counts} & \rot{CLEVR Distance} & \rot{Country211} & \rot{Describable Textures} & \rot{EuroSAT} & \rot{FGVC Aircraft} & \rot{Food-101} \\ \toprule
    ProLIP (pre-trained) & 
    74.56 & 96.42 & 83.25 & 29.78 & 15.13 & 21.29 & 66.91 & 60.89 & 38.02 & 91.03 \\ \midrule
    LongProLIP (S24M) & 70.76  \np{-3.80} &  95.64  \np{-0.78} &  81.17  \np{-2.08} &  32.18  \pp{2.40} &  15.79  \pp{0.66} &  18.89  \np{-2.40} &  65.37  \np{-1.54} &  65.24  \pp{4.35} &  34.39  \np{-3.63} &  87.71 \np{-3.32} \\
    LongProLIP (S128M) & 69.54  \np{-5.02} &  91.82  \np{-4.60} &  75.10  \np{-8.15} &  27.46  \np{-2.32} &  15.80  \pp{0.67} &  16.95  \np{-4.34} &  63.94  \np{-2.97} &  53.96  \np{-6.93} &  33.69  \np{-4.33} &  86.14 \np{-4.89} \\
    LongProLIP (SHD128M) & 74.58  \pp{0.02} &  96.47  \pp{0.05} &  82.94  \np{-0.31} &  31.70  \pp{1.92} &  15.80  \pp{0.67} &  21.29  \pp{0.00} &  66.33  \np{-0.58} &  57.94  \np{-2.95} &  37.57  \np{-0.45} &  90.74 \np{-0.29}
    \end{tabular}
    }
    \resizebox{\textwidth}{!}{
    \begin{tabular}{l*{11}{c}}
    \toprule
    & \rot{GTSRB} & \rot{Caltech-101} & \rot{ImageNet Sketch} & \rot{ImageNet v2} & \rot{ImageNet-A} & \rot{ImageNet-O} & \rot{ImageNet-R} & \rot{KITTI Vehicle Distance} & \rot{MNIST} & \rot{ObjectNet} \\ \toprule
    ProLIP (pre-trained) & 52.79 & 93.61 & 63.65 & 66.66 & 50.25 & 45.40 & 86.00 & 32.21 & 84.47 & 65.80 \\ \midrule
    LongProLIP (S24M) & 53.62  \pp{0.83} &  92.73  \np{-0.88} &  61.12  \np{-2.53} &  62.61  \np{-4.05} &  44.19  \np{-6.06} &  45.75  \pp{0.35} &  84.02  \np{-1.98} &  15.89  \np{-16.32} &  81.60  \np{-2.87} &  56.77 \np{-9.03} \\
    LongProLIP (S128M) & 48.97  \np{-3.82} &  92.88  \np{-0.73} &  60.30  \np{-3.35} &  61.31  \np{-5.35} &  41.91  \np{-8.34} &  46.15  \pp{0.75} &  83.17  \np{-2.83} &  21.52  \np{-10.69} &  82.34  \np{-2.13} &  55.72 \np{-10.08} \\
    LongProLIP (SHD128M) & 51.20  \np{-1.59} &  93.52  \np{-0.09} &  63.61  \np{-0.04} &  66.39  \np{-0.27} &  49.71  \np{-0.54} &  44.20  \np{-1.20} &  85.84  \np{-0.16} &  32.63  \pp{0.42} &  87.64  \pp{3.17} &  65.39 \np{-0.41} \\
    \end{tabular}
    }
    \resizebox{\textwidth}{!}{
    \begin{tabular}{l*{11}{c}}
    \toprule
 & \rot{Oxford Flowers-102} & \rot{Oxford-IIIT Pet} & \rot{Pascal VOC 2007} & \rot{PatchCamelyon} & \rot{Rendered SST2} & \rot{RESISC45} & \rot{Stanford Cars} & \rot{STL-10} & \rot{SUN397} & \rot{SVHN} \\ \toprule
ProLIP (pre-trained) & 78.37 & 93.45 & 81.74 & 61.41 & 54.31 & 68.27 & 91.33 & 97.91 & 71.35 & 72.77 \\ \midrule
LongProLIP (S24M) & 74.30  \np{-4.07} &  90.59  \np{-2.86} &  77.83  \np{-3.91} &  58.08  \np{-3.33} &  50.85  \np{-3.46} &  62.90  \np{-5.37} &  88.02  \np{-3.31} &  97.56  \np{-0.35} &  70.81  \np{-0.54} &  67.17 \np{-5.60} \\
LongProLIP (S128M) & 73.34  \np{-5.03} &  89.55  \np{-3.90} &  70.11  \np{-11.63} &  55.59  \np{-5.82} &  52.66  \np{-1.65} &  59.86  \np{-8.41} &  87.46  \np{-3.87} &  97.04  \np{-0.87} &  66.08  \np{-5.27} &  63.53 \np{-9.24} \\
LongProLIP (SHD128M) & 78.16  \np{-0.21} &  92.91  \np{-0.54} &  81.55  \np{-0.19} &  57.96  \np{-3.45} &  54.86  \pp{0.55} &  68.48  \pp{0.21} &  90.95  \np{-0.38} &  98.11  \pp{0.20} &  71.33  \np{-0.02} &  69.38 \np{-3.39} \\
\end{tabular}
}
\resizebox{\textwidth}{!}{
\begin{tabular}{l*{9}{c}}
\toprule
 & \rot{iWildCam} & \rot{Camelyon17} & \rot{FMoW} & \rot{Flickr} & \rot{MSCOCO} & \rot{WinoGAViL} & \rot{Dollar Street} & \rot{GeoDE} & \rot{Average} \\ \toprule
ProLIP (pre-trained) & 12.64 & 57.85 & 15.12 & 79.97 & 53.18 & 45.56 & 62.27 & 90.31 & 63.31 \\ \midrule
LongProLIP (S24M) & 13.69  \pp{1.05} &  50.65  \np{-7.20} &  0.00  \np{-15.12} &  79.44  \np{-0.53} &  55.88  \pp{2.70} &  48.61  \pp{3.05} &  57.24  \np{-5.03} &  89.02  \np{-1.29} &  60.48 \np{-2.83} \\
LongProLIP (S128M) & 12.37  \np{-0.27} &  51.63  \np{-6.22} &  0.00  \np{-15.12} &  78.20  \np{-1.77} &  55.12  \pp{1.94} &  45.59  \pp{0.03} &  56.19  \np{-6.08} &  86.47  \np{-3.84} &  58.67 \np{-4.64} \\
LongProLIP (SHD128M) & 13.54  \pp{0.90} &  60.27  \pp{2.42} &  14.54  \np{-0.58} &  80.13  \pp{0.16} &  53.54  \pp{0.36} &  51.98  \pp{6.42} &  63.20  \pp{0.93} &  90.61  \pp{0.30} &  63.34 \pp{0.03} \\
\bottomrule
\end{tabular}
}
\end{table}

\end{document}